\documentclass[runningheads]{llncs}
\usepackage{graphicx}
\usepackage{comment}
\usepackage{amsmath,amssymb} 
\usepackage{color}
\usepackage{bm}
\usepackage{wrapfig}
\usepackage{subfigure}
\usepackage{booktabs}
\usepackage{tabularx}
\usepackage{multirow}
\def\etal{\emph{et al.~}}
\def\ie{\emph{i.e.}}
\def\eg{\emph{e.g.}}

\newcommand\blfootnote[1]{%
  \begingroup
  \renewcommand\thefootnote{}\footnote{#1}%
  \addtocounter{footnote}{-1}%
  \endgroup
}

\begin{document}
\pagestyle{headings}
\mainmatter

\title{Hierarchical Kinematic Human
Mesh Recovery}

\titlerunning{HKMR}

\author{Georgios Georgakis*\inst{1,2} \and Ren Li*\inst{1} \and Srikrishna Karanam \inst{1} \and Terrence Chen \inst{1} \and Jana Ko{\v{s}}eck{\'a} \inst{2} \and Ziyan Wu \inst{1}}

\authorrunning{G. Georgakis, R. Li, S. Karanam, T. Chen, J. Kosecka, and Z. Wu}

\institute{United Imaging Intelligence, Cambridge MA, USA \and
 George Mason University, Fairfax VA, USA \\
\email{\{first.last\}@united-imaging.com}}

\maketitle

\begin{abstract}
We consider the problem of estimating a parametric model of 3D human mesh from a single image. While there has been substantial recent progress in this area with direct regression of model parameters, these methods only implicitly exploit the human body kinematic structure, leading to sub-optimal use of the model prior. In this work, we address this gap by proposing a new technique for regression of human parametric model that is explicitly informed by the known hierarchical structure, including joint interdependencies of the model. This results in a strong prior-informed design of the regressor architecture and an associated hierarchical optimization that is flexible to be used in conjunction with the current standard frameworks for 3D human mesh recovery. We demonstrate these aspects by means of extensive experiments on standard benchmark datasets, showing how our proposed new design outperforms several existing and popular methods, establishing new state-of-the-art results. By considering joint interdependencies, our method is equipped to infer joints even under data corruptions, which we demonstrate by conducting experiments under varying degrees of occlusion.
\end{abstract}

\section{Introduction}
\label{sec:intro}

\blfootnote{* Georgios Georgakis and Ren Li are joint first authors and contributed equally to this work done during their internships with United Imaging Intelligence, Cambridge MA, USA. Corresponding author: Srikrishna Karanam.}We consider 3D human mesh recovery, \ie, fitting a parametric model to an image of a person that best explains the body pose and shape. With a variety of applications \cite{singh2017darwin,martinez2018real}, there has been notable recent interest in this field \cite{bogo2016keep,tung2017self,kanazawa2018end,kolotouros2019learning}.

The dominant paradigm for this problem involves an \textit{encoder-regressor} architecture \cite{kanazawa2018end}; the deep CNN encoder takes the input image and produces the feature representation which the regressor processes to produce the model parameters. A recent method \cite{kolotouros2019learning} extends this to the \textit{encoder-regressor-optimizer} framework by introducing an in-the-loop optimization step \cite{bogo2016keep} which uses the output of the regressor as the starting point to iteratively optimize and generate more accurate model estimates. The regressor forms the core of both of these approaches and is typically realized with a block of fully-connected layers with non-linear activation units, taking feature vectors as input and producing the shape and rotation/pose parameter vectors as output. 

However, it has been shown \cite{kendall2017geometric} that direct regression of rotation parameters is a very challenging task. The difficulty is exacerbated in our case of human joints due to multiple rotations and their dependencies, as noted in prior work \cite{kanazawa2018end,kolotouros2019convolutional}. Kendall \etal \cite{kendall2017geometric} further notes such regression tasks can be made more amenable, with significant performance improvements, by grounding regressor design considerations in geometry which in our context is the underlying structure of the model we are attempting to fit. However, existing \textit{encoder-regressor} methods do not include such structural information in their design, leaving much room for performance improvement. This is even more pronounced in situations involving data corruptions (\eg, occlusions), where intuitively structural information (\eg, one joint dependent on or connected to another joint) can readily help infer these model parameters even when one or more joints are occluded. 

To this end, we present a new architecture, called \textbf{HKMR}, and an associated hierarchical optimization technique, for 3D human mesh recovery. While Kolotouros \etal \cite{kolotouros2019convolutional} avoids direct regression of model parameters and instead estimates the 3D mesh vertices, we investigate a more model-structure-informed design that exploits the strengths of such a representation. We use the popular SMPL model \cite{loper2015smpl}, which is based on the standard skeletal rig with a well-known hierarchical structure of chains with interdependent joints. Note however that HKMR can be used with other hierarchical human model instantiations as well.

\begin{figure}[t]
\centering
\includegraphics[width=1\linewidth]{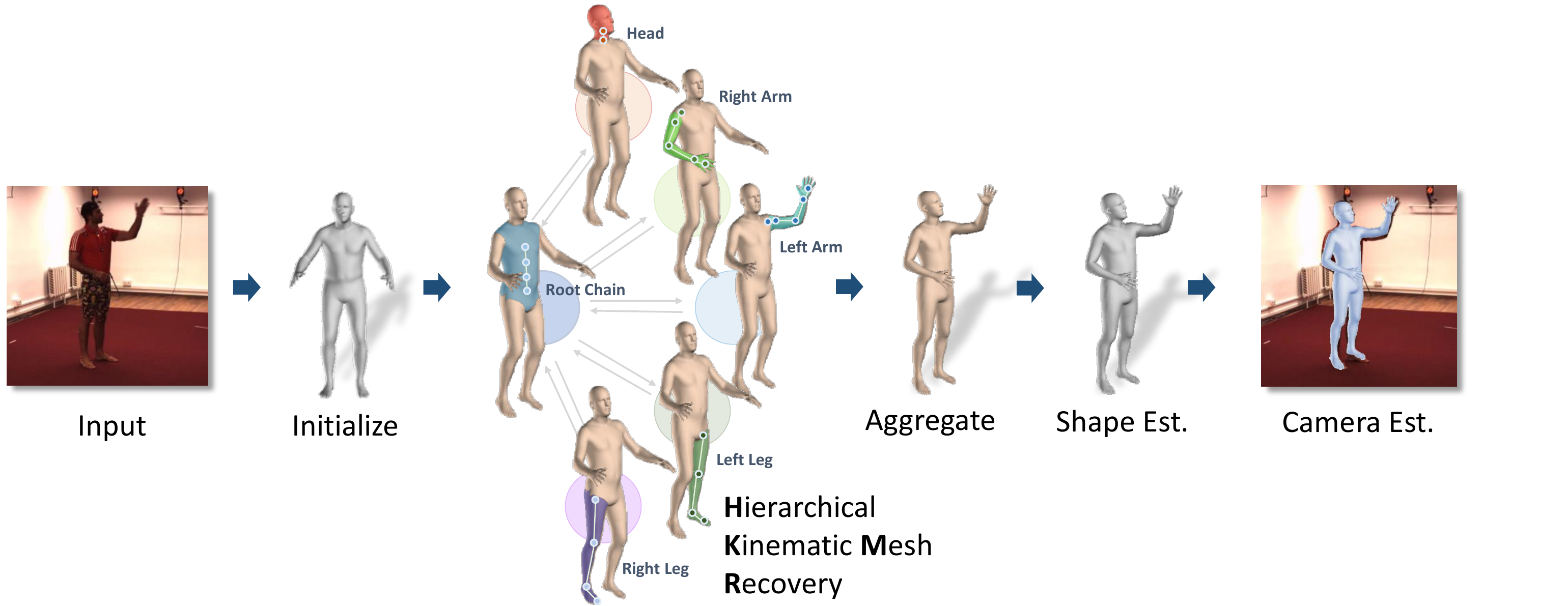}
\caption{We present HKMR, a new structure-informed design that hierarchically regresses the pose and shape parameters of the human body mesh.}
\label{fig:teaser}
\end{figure}

\begin{figure}[t]
\centering
\includegraphics[width=0.9\linewidth]{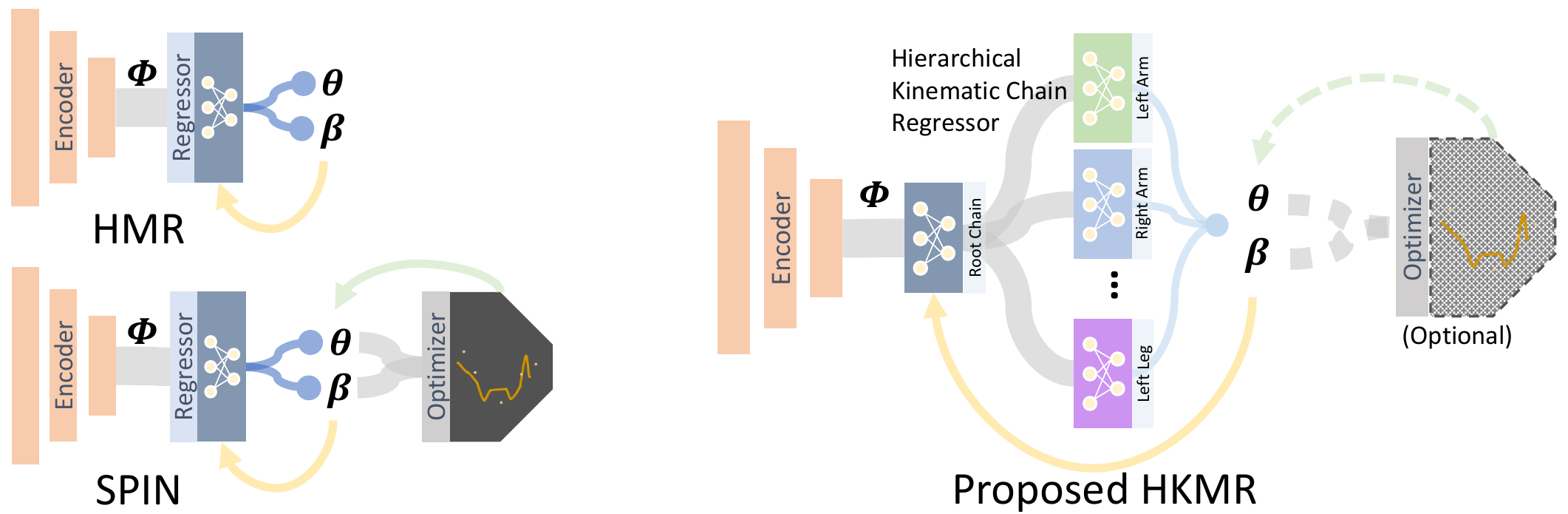}
\caption{HKMR can be used in either the \textit{encoder-regressor} paradigm (HMR \cite{kanazawa2018end}) or the \textit{encoder-regressor-optimizer} paradigm (SPIN \cite{kolotouros2019learning}).}
\label{fig:block_compare}
\end{figure}

HKMR defines six chains following the standard skeletal rig (a root chain and five dependent child chains: head, left/right arms, left/right legs). Each chain is designed following the skeletal rig's kinematic model, with explicit interdependencies of joints. We repeat this for all the chains, with each non-root chain's predictions conditioned on the root chain's output, thus modeling HKMR with a set of hierarchically-nested operations (see Figure~\ref{fig:teaser}). As one can note, this is in stark contrast to the existing paradigm~\cite{kanazawa2018end} that simply operates in a \textit{features-in-parameters-out} fashion without explicitly exploiting the underlying structure of the model. Furthermore, such a design for the regressor is particularly beneficial for parameter inference under data corruptions. By modeling hierarchical joint interdependencies, HKMR facilitates the inference of the current joint even if the previous joint is unreliable or unobserved due to occlusions or other reasons. We show HKMR leads to a new architecture for 3D human mesh estimation that substantially outperforms currently dominant baseline architectures for this problem \cite{kanazawa2018end,kolotouros2019convolutional}. Our method is flexible to be used in both the \textit{encoder-regressor} and \textit{encoder-regressor-optimizer} paradigms (see Figure~\ref{fig:block_compare}, where we show how the \textit{optimizer} can be optionally added to our pipeline), and we demonstrate substantial performance improvements in both these cases. 

To summarize, our key contributions include:
\begin{itemize}
    \item We present HKMR, a new parameter regressor that is flexible to be used in either of the two existing \textit{encoder-regressor} or \textit{encoder-regressor-optimizer} paradigms for 3D human mesh recovery.
    \item Our key insight is the design of the parameter regressor in a way that explicitly exploits structural constraints of the human body model, resulting in a strong model-design-informed architecture.
    \item We empirically show HKMR improves the performance of both \textit{encoder-regressor} and \textit{encoder-regressor-optimizer} methods and establishes new state-of-the-art results on several human mesh recovery benchmark datasets. 
    \item HKMR is robust to occlusions, as evidenced by substantial relative performance improvements on data under a wide variety of occlusion conditions.
\end{itemize}

\section{Related Work}
\label{sec:relatedWork}
There is a large body of work on human pose estimation, formulating the problem as one of
predicting 2D keypoints \cite{cao2018openpose,carreira2016human,alp2018densepose}, estimating 3D joints \cite{mehta2017vnect,moreno2017matrix,pavlakos2018ordinal}, or model-based parametric human body estimation \cite{guler2019holopose,zhang2019danet,xu2019denserac,kocabas2019vibe,arnab2019exploiting,pavlakos2019texturepose}. Here, we discuss most relevant methods with particular focus on their structure-related design choices. \\
\indent \textbf{Flat-regression methods.} A common paradigm has been the direct CNN-based regression of the body model parameters with focus on capturing multi-scale information \cite{newell2016stacked,pavlakos2017coarse} or encoding spatial relationships between keypoints \cite{wei2016convolutional,toshev2014deeppose,martinez2017simple}. With the increasing adoption of parametric human models such as SMPL~\cite{loper2015smpl}, several methods shifted focus to regressing the model parameters \cite{tan2018indirect,kanazawa2018end,pavlakos2018learning,omran2018neural,yao2019densebody,varol2018bodynet,lassner2017unite}. The most representative in this line of work is of Kanazawa \etal\cite{kanazawa2018end} that learned an end-to-end image-to-SMPL-parameter regressor. Kolotouros \etal\cite{kolotouros2019learning} extended this work by combining the initial regressor estimates with an in-the-loop SMPLify \cite{bogo2016keep} optimization step, reporting the most competitive performance on benchmark datasets to date. However, these methods tend to ignore the inherent structure of the body model in their regressor design. \\
\indent \textbf{Structure-aware methods.}
Several methods have sought to include structural priors in their design.
Earlier works considered pictorial structures~\cite{felzenszwalb2005pictorial,yang2012articulated}
where the human body was modeled as a set of rigid templates with pairwise potentials. More recently, Cai \etal\cite{cai2019exploiting} and Kolotouros \etal\cite{kolotouros2019convolutional} exploited variants of graph CNNs to encode the 2D keypoint or mesh structure. Ci \etal\cite{ci2019optimizing} went a step further and proposed a generic framework for graph CNNs, introducing a specific formulation focusing on local joint neighborhoods. Aksan \etal\cite{aksan2019structured} explicitly modeled the joint dependencies by means of a structured prediction layer. Tang \etal\cite{tang2019does} investigated the relationship between different parts of the human body to learn part-specific features for 2D keypoint prediction. Fang \etal\cite{fang2018learning} encoded human body dependencies by employing Bi-directional RNNs during 3D joint prediction, while Isack \etal\cite{isack2020repose} attempted to learn priors between keypoints by following a pre-specified prediction order and connectivity. In contrast to these approaches, our method explicitly encodes both the hierarchical structure as well as joint interdependencies through a kinematics model. Zhou \etal \cite{zhou2016deep} also proposed kinematic modeling but is substantially different from our method. First, it does not model joint interdependencies in each chain, thus failing to take full advantage of the kinematic formulation. Next, it models all joints as part of one large chain, therefore not exploiting the skeletal rig's hierarchical nature. Finally, it only generates 3D joints and not the full mesh.

\section{Approach}

As noted in Sections~\ref{sec:intro} and~\ref{sec:relatedWork}, existing mesh recovery methods formulate the problem as purely one of regressing the real-valued parameter vectors. While the network (during training) is regularized by priors learned from data (mixture models as in \cite{bogo2016keep}) or even discriminator models \cite{kanazawa2018end}, we contend this does not fully exploit the knowledge we have about the structure of the SMPL model. From SMPL's design principles, we know it is motivated by the standard skeletal rig, and that this rig has an associated hierarchy. We argue that a parameter estimation procedure that is explicitly informed by this hierarchy constitutes a stronger integration of the structure of the SMPL model as opposed to HMR-like \cite{kanazawa2018end} methods that generate the parameter vectors by means of a structure-agnostic set of fully-connected units (see HMR in Figure~\ref{fig:block_compare}). Motivated by this intuition, we propose \textbf{HKMR}, a new architecture for the parameter regressor. As we demonstrate in Section~\ref{sec:expResults}, this leads to a new convolutional neural network architecture that substantially outperforms the currently dominant paradigm of \textit{encoder-regressor} architectures \cite{kanazawa2018end,kolotouros2019convolutional}, while also lending itself favorably applicable to \textit{encoder-regressor-optimizer} approaches like SPIN \cite{kolotouros2019learning}.

\subsection{3D Body Representation}
We use the SMPL model of Loper \etal \cite{loper2015smpl} to parameterize the 3D human mesh. SMPL is a differentiable model defined in the real-valued space of pose $\bm{\Theta} \in \mathbb{R}^{72}$ and shape $\bm{\beta} \in \mathbb{R}^{10}$ parameters.  While $\bm{\Theta}$ models the relative 3D rotation of $K=24$ joints in the axis-angle representation, $\bm{\beta}$ models the shape of the body as captured by the first 10 coefficients of a PCA projection of the shape space. SMPL defines a function $\mathcal{M}(\bm{\Theta}, \bm{\beta}) \in \mathbb{R}^{N \times 3}$ that produces the $N=6890$ 3D vertices representing the human mesh. Starting from a template mesh, the desired body mesh is obtained by applying forward kinematics based on the joint rotations $\bm{\Theta}$ and by applying shape deformations conditioned on both $\bm{\Theta}$ and $\bm{\beta}$. Finally, the joint locations are defined as a linear combination of the mesh vertices, obtained by a linear regression function $\mathcal{X} (\mathcal{M}(\bm{\Theta}, \bm{\beta})) \in \mathbb{R}^{K \times 3}$.

\begin{figure}[t]
\centering
\subfigure[Chains and joints]{\label{fig:chains}
\includegraphics[width=4.8cm,height=4.2cm]{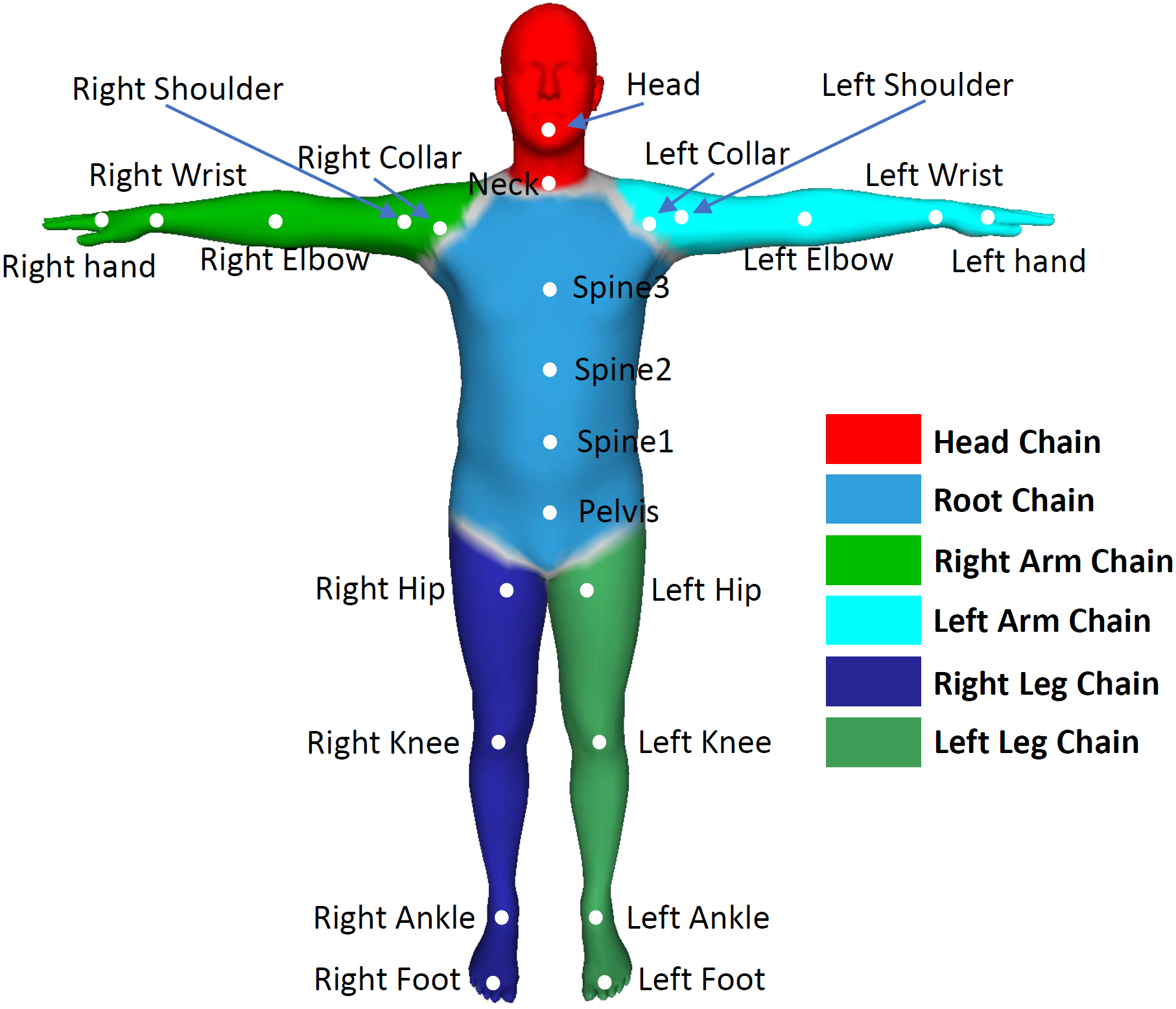}}
\subfigure[Parameter estimation]{\label{fig:estimation}
\includegraphics[width=7cm]{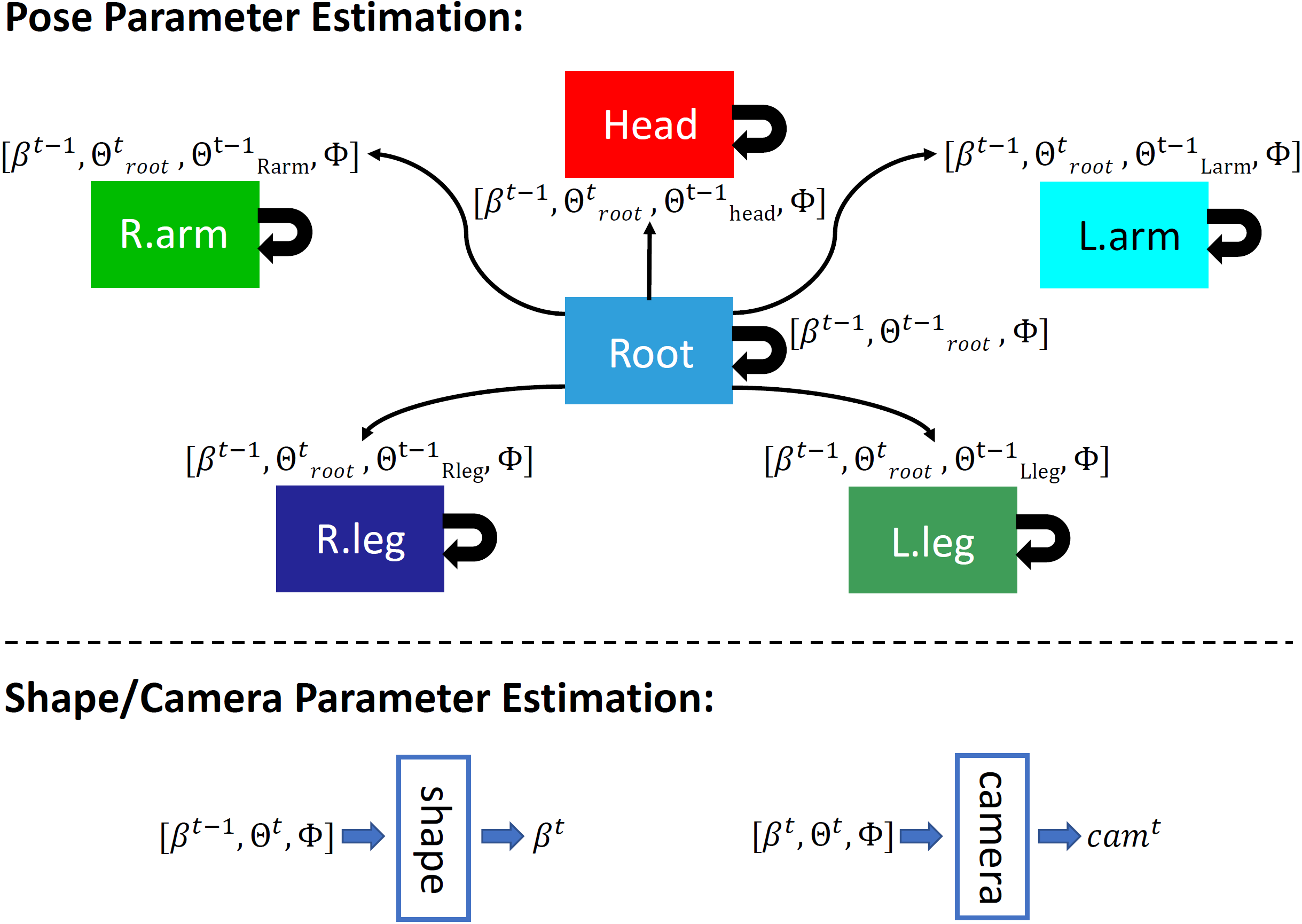}}
\caption{The kinematic chains and our hierarchical optimization workflow.}
\label{fig:chains_joints}
\end{figure}

\subsection{Hierarchical Kinematic Pose and Shape Estimation}

Our architecture comprises an encoder to generate features (the same ResNet50 \cite{he2016deep} to existing models \cite{kanazawa2018end,kolotouros2019convolutional}), followed by our proposed regressor that explicitly models the hierarchy between various body parts and the interdependency of the body joints within these parts. We define each body part by a standalone kinematic chain with joints at the same locations as the SMPL model (see Figure~\ref{fig:chains}). As in the standard skeletal rig, we consider the chain representing the torso of the body as root, with all other chains (arms, leg, and head) hierarchically dependent on this root chain. We estimate $\bm{\Theta}$ and $\bm{\beta}$ in the same spirit as the iterative feedback of HMR \cite{kanazawa2018end} but is realized substantially differently. Given six chains corresponding to different parts of the body, we express $\bm{\Theta}$ as a concatenation of all the individual chain pose parameters: $\bm{\Theta}=[\bm{\theta}_{\text{root}},\bm{\theta}_{\text{head}},\bm{\theta}_{\text{R.arm}},\bm{\theta}_{\text{L.arm}},\bm{\theta}_{\text{R.leg}},\bm{\theta}_{\text{L.leg}}]$, where each $\bm{\theta}$ represents chains' angle-axis pose parameters. For example root chain is comprised by 4 joints, $\bm{\theta}_{root}$ is of dimensionality 12. Starting from the mean pose and shape $\bm{\Theta}^0$ and $\bm{\beta}^0$, our method first estimates the next $\bm{\Theta}^t$, which is then used to update the next $\bm{\beta}^t$. This process is repeated for $t=T$ iterations resulting in final estimates $\hat{\bm{\Theta}}=[\hat{\bm{\theta}}_{\text{root}},\hat{\bm{\theta}}_{\text{head}},\hat{\bm{\theta}}_{\text{R.arm}},\hat{\bm{\theta}}_{\text{L.arm}},\hat{\bm{\theta}}_{\text{R.leg}},\hat{\bm{\theta}}_{\text{L.leg}}]$ and $\hat{\bm{\beta}}$. This constitutes what we call the \textbf{outer iterative} process involving all the chains. Each chain $c$ also has an \textbf{inner iterative} process estimating its pose $\hat{\bm{\theta}}_{c}^{t}$ at each outer iterative step $t$, \ie, $\hat{\bm{\theta}}_{c}^{t}$ itself is updated for multiple iterations at the outer step $t$. In the following, we first describe the inner iterative process of each chain $c$, followed interaction between inner and outer iterative steps, leading up to the overall hierarchical process for parameter optimization.

\paragraph{\textup{\textbf{Iterative kinematic chain parameter estimation.}}} For simplicity of exposition, we focus on one chain here. As noted above, at each outer iterative step $t$, the chain $c$ has an inner-iterative process to estimate its pose parameters $\bm{\theta}^{t}$. Specifically, the chain $c$ takes as input its previous estimates at $t-1$, $\bm{\theta}^{t-1}=\bm{\theta}^I$ and $\bm{\beta}^{t-1}=\bm{\beta}^I$ and iteratively refines it yielding $\bm{\theta}^{t}$ at the current outer step $t$. In the following section we drop the superscripts $t-1$ and $t$ for clarity. 

We take inspiration from inverse kinematics, more specifically from iterative solvers for the problem. The 3D location of the chain's end-effector $e = [e_x, e_y, e_z]^T$ is related to the pose of the rigid bodies in the chain by a nonlinear function characterizing forward kinematics $\bm{e} = g(\bm{\theta})$. For inverse kinematics, we seek $\bm{\theta}$ (system configuration) that will realize it: $\bm{\theta} = g^{-1}(\bm{e})$. Considering the Taylor series expansion of $g$, we can characterize changes in the end-effector's current position $e$ relative to changes in $\bm{\theta}$ in terms of the Jacobian matrix $\bm{J}(\bm{\theta})$ of partial derivatives of the system as $\Delta\bm{e} = \bm{J} \Delta \bm{\theta}$. Since we are interested in the inverse estimation (\ie, how $\bm{\theta}$ changes with respect to $\bm{e}$), the pseudo-inverse of the Jacobian $\bm{J}^+$ is used to estimate the residual $\Delta\bm{\theta} = \bm{J}^+ \Delta \bm{e}$, followed by the update $\bm{\theta} \leftarrow \bm{\theta} + \alpha\Delta\bm{\theta}$, where $\alpha$ is a small scalar. This update is repeated in the inner-iterative process until a proximity to the goal end-effector position criterion is reached. This is the essence of iterative solvers for inverse kinematics problems frequently used for kinematic chains in robotics. With the forward kinematics model being a continuous function of $\bm{\theta}$ \cite{zhou2016deep}, we can incorporate this  solution framework into an end-to-end learning paradigm. In other words, we design a learnable function for the chain $c$ that predicts the residuals $\Delta\bm{\theta}$ and updates $\bm{\theta}$ iteratively with the available 3D joint annotations as supervision.

\begin{figure}[t]
\centering
\includegraphics[width=0.8\linewidth]{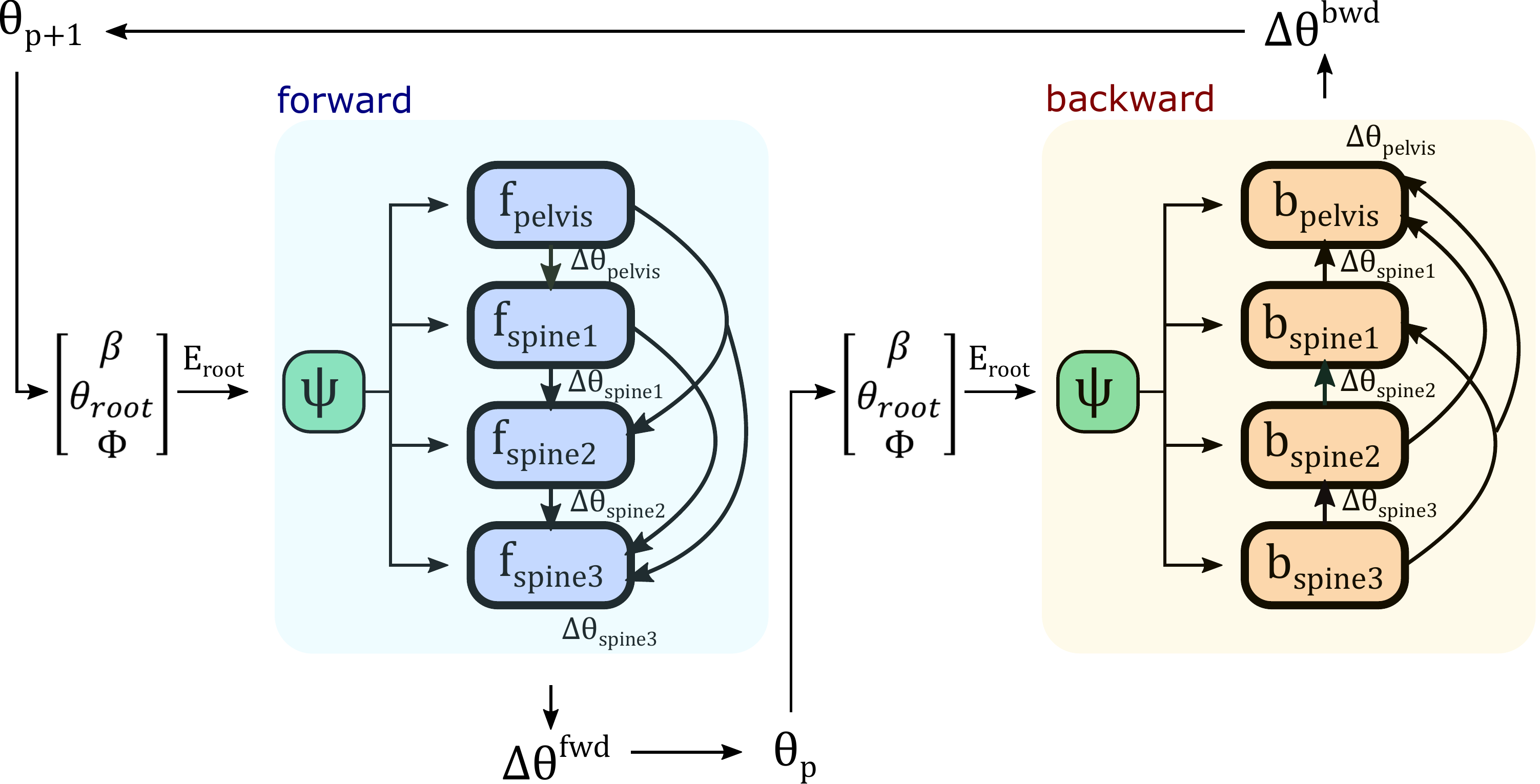}
\caption{The forward-backward cycle in $Q_{root}$ that corresponds to the inner iterative procedure of each chain (here we use the root chain as example). $E_{root}$ produces a common embedding $\psi$ between the model parameters $\theta$, $\beta$ and image features $\Phi$. Note $\theta_{root}$ is replaced by $\theta_p$/$\theta_{p+1}$ at the end of the forward/backward process respectively.}
\label{fig:circle}
\end{figure}

Specifically, we estimate the $\bm{\theta}$ for each kinematic chain $c$ with a trainable neural network $Q_c$ that takes as input values $\bm{\beta}^I$, $\bm{\theta}^I$, and image features $\bm{\Phi}$ extracted from the encoder. Given $\bm{\Phi}$, $\bm{\beta}^I$ and the chain-specific $\bm{\theta}^I$, we first learn a low-dimensional embedding $\bm{\psi} \in \mathbb{R}^{d}$ with $Q_c$'s embedding module $E_c$. Our key insight is that the predicted angle of a certain joint affects the predictions of all the following joint angles in the chain. Any predicted angle in the chain changes the system configuration, consequently requiring the adjustment of the following angle predictions, which we do iteratively. To this end, we predict the $\Delta\bm{\theta}_{i,p}$ of the $i^{th}$ joint ($i=0,1,\cdots,K_c-1$, $K_c$: number of joints in $c$) starting from $i=0$ at the inner iteration step $p$ through a \textit{forward} pass of the kinematic chain. In this process, we concatenate $\bm{\psi}$ and the estimated residuals of all previous joints in the following chain sequence:
\begin{equation}\label{eq:delta_theta_forward}
    \Delta\bm{\theta}_{i,p}^{\text{fwd}} = f_i([\bm{\psi}, \Delta\bm{\theta}_{i-1,p}, \Delta\bm{\theta}_{i-2,p},\cdots, \Delta\bm{\theta}_{0,p}]).
\end{equation}
where the function $f_i$ is realized as one fully-connected layer for the joint $i$ that outputs three real-valued numbers. See Figure~\ref{fig:circle} for a visual summary. For the first joint $i=0$ (\texttt{pelvis}) in the chain, we use $[\bm{\psi}, 0]$ as input to $f_i$. In other words, the second joint's prediction is dependent on the first (\texttt{spine1} depends on \texttt{pelvis}), the third is dependent on both first and second (\texttt{spine2} depends on both \texttt{spine1} and \texttt{pelvis}), and so on. When all residuals are predicted, the current estimate (at inner iteration step $p$) for the pose of the chain $\bm{\theta}_{p}=[\bm{\theta}_{0,p},\cdots,\bm{\theta}_{K_c-1,p}]$ is updated as $\bm{\theta}_{i,p}=\bm{\theta}_{i,p-1}+\Delta\bm{\theta}_{i,p}^{\text{fwd}}$; the embedding $\bm{\psi}$ is then updated using $E_c$ based on this new $\bm{\theta}_{p}$. Specifically, as above, this step takes $\bm{\Phi}$, $\bm{\beta}^I$ and the updated $\bm{\theta}_{p}$ as input, producing an updated $\bm{\psi}$. 
Since these joint angles can be affected by both the next and previous joints, after the forward update as above, we additionally perform a \textit{backward} pass: 
\begin{equation}\label{eq:delta_theta_backward}
    \Delta\bm{\theta}_{i,p+1}^{\text{bwd}} = b_i([\bm{\psi}, \Delta\bm{\theta}_{i-1,p+1}, \Delta\bm{\theta}_{i-2,p+1},\cdots, \Delta\bm{\theta}_{0,p+1}]) 
\end{equation}
where $b_i$ is defined similarly to $f_i$ above.
Note that the notation $p$ is used to differentiate between the forward and backward pass (i.e. $\theta_p$ referred to the estimated pose parameters after the forward pass, and $\theta_{p+1}$ to the corresponding backward).
The backward update takes the updated embedding $\bm{\psi}$ from the forward update as the input and starts from the last joint $i=K_c-1$ in the forward update (\texttt{spine3}, see backward in Figure~\ref{fig:circle}). Specifically, in predicting the $\Delta\bm{\theta}$ for \texttt{spine3}, the input to its corresponding $b_i$ is $[\bm{\psi}, 0]$, as with the initial step in the forward update as above. Subsequently, every other joint prediction depends on all the preceding predictions (\eg, \texttt{spine2} depends on \texttt{spine3}, \texttt{spine1} depends on both \texttt{spine2} and \texttt{spine3}, and so on; see backward in Figure~\ref{fig:circle}). Once the backward residuals are computed, the current estimate for the pose of the chain $\bm{\theta}_{p+1}=[\bm{\theta}_{0,p+1},\cdots,\bm{\theta}_{K_c-1,p+1}]$ at inner iteration step $p+1$ (the forward pass above is step $p$) is updated as $\bm{\theta}_{i,p+1}=\bm{\theta}_{i,p}+\Delta\bm{\theta}_{i,p+1}^{\text{bwd}}$. Again, as above, given this updated $\bm{\theta}_{p+1}$, we update $\bm{\psi}$ using $E_c$. This forward-backward cycle (forward is inner iterative step $p$, backward is inner iterative step $p+1$) is repeated for multiple iterations. Following the inverse kinematics formulation, we seek to optimize the prediction of $\bm{\theta}$ so we can reach the position of the chain's end-effector $e$ given the desired pose. This constitutes defining an $e$ in the chain and using an $L_1$-like distance function to minimize the distance between $e$ and its ground truth. In practice, since inverse kinematics can produce multiple solutions, we apply this loss on all joints in the chain, as we show next.

\paragraph{\textup{\textbf{Hierarchical Optimization.}}} As noted above, we denote the torso kinematic chain as root, and the others as its $n_c$ dependent/children chains. The purpose of having this hierarchy is to allow the pose predicted by the root chain to affect how the rest of the chains operate. This is achieved by using the prediction of the root chain's $\bm{\theta}$ as part of the input to all other chains' neural networks. Specifically, while the root chain network takes $\bm{\beta}^{t-1}$, $\bm{\theta}_{\text{root}}^{t-1}$, and $\Phi$ as input in predicting the next $\bm{\theta}_{\text{root}}^{t}$, all other chains' networks take the previous $\bm{\beta}^{t-1}$, $\bm{\theta}_{\text{c}}^{t-1}$, $\Phi$, and the current $\bm{\theta}_{\text{root}}^{t}$ as input in predicting the next $\bm{\theta}_{\text{c}}^{t}$. This is also visually summarized in Figure~\ref{fig:chains_joints}(b). Since the kinematic chains operate only on the pose parameters, the shape parameters $\bm{\beta}$ remain constant during this process. We re-estimate $\bm{\beta}$ after every $\bm{\Theta}$ prediction cycle ends (\ie, after each of the six chains have completed their inner-iterative estimation process).
To this end, we define a shape-estimation neural network that takes the previous outer iteration step's shape prediction $\hat{\bm{\beta}}^{t-1}$, the current outer iteration's pose prediction $\hat{\bm{\Theta}}^{t}$, and the features $\Phi$ as input and produces the current outer iteration step's shape estimate $\hat{\bm{\beta}}^{t}$. This updated $\hat{\bm{\beta}}^{t}$ (along with $\hat{\bm{\Theta}}^{t}$) will then be used to initialize the next (outer iteration step $t+1$) $\bm{\Theta}$ prediction cycle.
We supervise the prediction of both pose and shape parameters by applying an $L_1$ loss between the predicted 3D joint locations $\hat{\bm{X}}^{t}=\mathcal{X} (\mathcal{M}(\hat{\bm{\Theta}}^{t},\hat{\bm{\beta}}^{t}))$ of the chain and their respective ground-truth:
$\label{eq:joints_3d}
    L_{\text{3D}}^{t} = \sum_{i=1}^{N_{\text{3d}}} ||\hat{\bm{X}}_i^{t} - \bm{X}_i||_1
$, where the subscript $i$ represents the $i^{th}$ joint and $N_{\text{3d}}$ is the number of available annotated 3D joints. Note that if SMPL parameter ground-truth is available, we can also use this to directly supervise the pose and shape parameters:
$L_{\text{smpl}}^{t} = ||[\hat{\bm{\Theta}}^{t}, \hat{\bm{\beta}}^{t}] - [\bm{\Theta}, \bm{\beta}]||^2_2$.

\paragraph{\textup{\textbf{Camera Parameter Estimation.}}} In order to fully utilize the 2D joints annotations available in most datasets, we define a camera-estimation network that takes the predicted parameters $\hat{\bm{\Theta}}^{t}$, $\hat{\bm{\beta}}^{t}$ at the outer step $t$, and image features $\bm{\Phi}$ to estimate the camera parameters that model a weak-perspective projection as in HMR \cite{kanazawa2018end}, giving translation $\bm{\rho^{t}} \in \mathbb{R}^2$ and scale $s^{t}\in \mathbb{R}$. Consequently, 2D joints $\hat{\bm{x}}$ can be derived from the 3D joints $\hat{\bm{X}}^{t}$ as $ \hat{\bm{x}}^{t}=s\Pi(\hat{\bm{X}}^{t}) + \bm{\rho}^{t} $, where $\Pi$ is an orthographic projection. We then supervise the estimated 2D joints with an L1 loss:
$L_{\text{2D}}^{t} = \sum_{i=1}^{N_{\text{2d}}}||\hat{\bm{x}}_{i}^{t} - \bm{x}_{i}||_1$, where the subscript $i$ represents the $i^{th}$ joint and $N_{\text{2d}}$ is the number of available annotated 2D joints.

\paragraph{\textup{\textbf{Pose Prior.}}} To ensure realism, we follow Pavlakos \etal \cite{pavlakos2019expressive} and train a VAE to learn a distribution over plausible human poses. 
The VAE encodes the 69-D $\bm{\Theta}$ (corresponding to 23 joints excluding the global orientation) to the latent vector $\bm{Z}_{\bm{\Theta}}$, which is then used, via re-parameterization \cite{kingma2013auto}, by the decoder to reconstruct $\bm{\Theta}$. 
We use the MoSh dataset \cite{loper2014mosh}, comprising about 6 million synthetic human poses and shapes, to train our VAE with the standard learning objective. Once trained, we discard the decoder and only use the encoder to ensure the pose predicted by our regressor is physically plausible. To this end, we use this encoder to compute the latent representation $\bm{Z}_{\hat{\bm{\Theta}}^t}$ of the predicted $\hat{\bm{\Theta}}^t$ at the current outer iteration step $t$. We then enforce $\bm{Z}_{\hat{\bm{\Theta}}^t}$ to follow the same unit normal distribution used to train the encoder, which is realized with the KL divergence loss at outer iteration step $t$: $L_{\text{KL}}^t = KL(\bm{Z}_{\hat{\bm{\Theta}}^t}||\mathcal{N}(\bm{0},\mathcal{\bm{I}}))$.

\subsection{Overall Learning Objective}
During training, we add up all the losses over the T outer iterations ($\lambda$s are the corresponding loss weights):
\begin{equation}\label{eq:overall_loss}
    L = \sum_{t=1}^T\lambda_{\text{smpl}} L_{\text{smpl}}^{t} + \lambda_{\text{3D}} L_{\text{3D}}^{t} + \lambda_{\text{2D}} L_{\text{2D}}^{t} + \lambda_{\text{KL}} L_{\text{KL}}^{t},
\end{equation}
With Equation~\ref{eq:overall_loss}, we perform a single backward pass during which the individual chain models, the shape model, and the camera model are optimized jointly.

\subsection{In-the-loop Optimization}
\label{sec:inTheLoopSec}
SPIN \cite{kolotouros2019learning} introduced the \textit{encoder-regressor-optimizer} paradigm with an in-the-loop SMPLify \cite{bogo2016keep} optimization step. Like \textit{encoder-regressor} above, our method can be used in this framework as well. Since SPIN \cite{kolotouros2019learning} used an HMR-inspired \cite{kanazawa2018end} regressor, our proposed regressor can be used as a direct drop-in replacement, and we show in Section~\ref{sec:expResults} this results in performance improvements.

\section{Experiments and Results}
\label{sec:expResults}
\paragraph{\textup{\textbf{Datasets.}}} Following HMR \cite{kanazawa2018end}, we use the training splits of LSP \cite{johnson2010clustered}, LSP-extended \cite{johnson2011learning}, MPII \cite{andriluka20142d}, MS COCO \cite{lin2014microsoft}, Human3.6M \cite{ionescu2013human3} and MPI-INF-3DHP \cite{mehta2017monocular} for network training. We use the same encoder as HMR \cite{kanazawa2018end} (ResNet50) and exactly follow their experimental setup, reporting results for both protocols P1 and P2 of Human3.6M (P1 uses videos of all cameras whereas P2 uses only the frontal one - camera 3). We use the mean per joint position error (MPJPE) without any Procrustes post-processing as our evaluation metric.  Additional implementation information results/discussion are in the supplementary material.

\begin{figure}
\centering
\subfigure[Original]{\label{fig:ori}
\includegraphics[height=2.3cm]{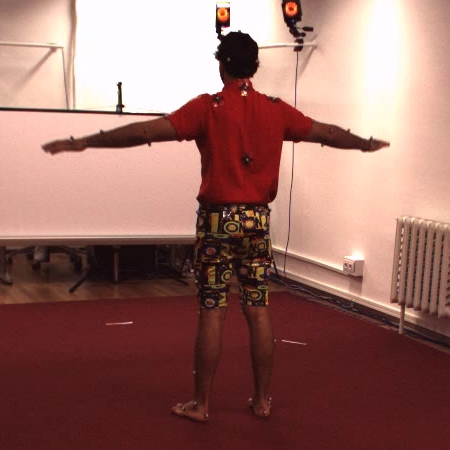}}
\subfigure[Oriented bar]{\label{fig:bar}
\includegraphics[height=2.3cm]{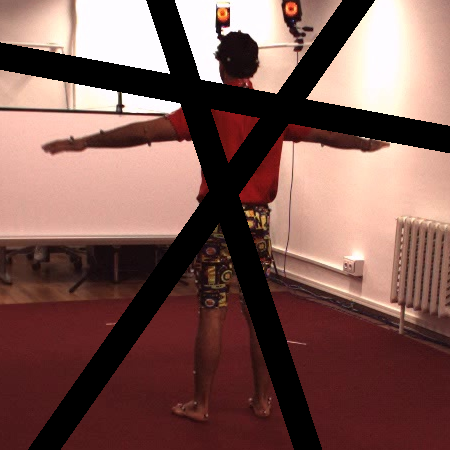}}
\subfigure[Circle]{\label{fig:cir}
\includegraphics[height=2.3cm]{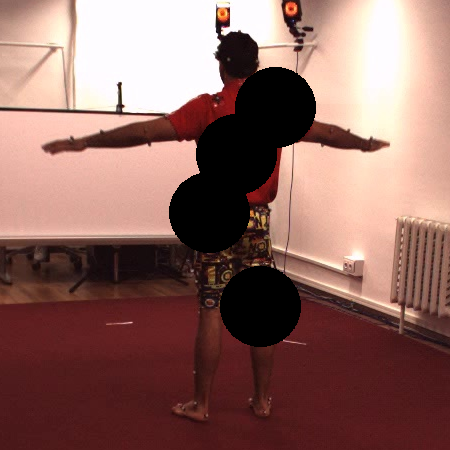}}
\subfigure[Rectangle]{\label{fig:rec}
\includegraphics[height=2.3cm]{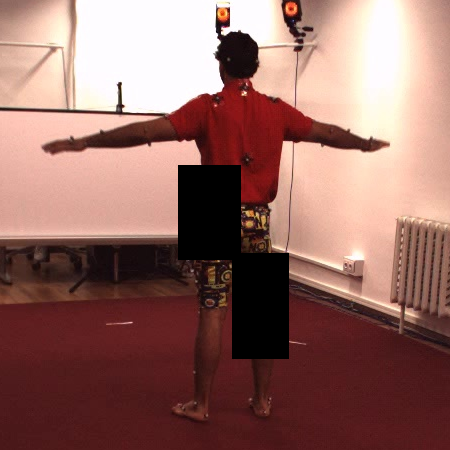}}
\caption{Examples of occlusion patterns.}
\label{fig:occlusion}
\end{figure}

\paragraph{\textup{\textbf{Occlusions.}}} In order to demonstrate robustness to occlusions, given an image, we generate multiple synthetically occluded images. Specifically, following Sarandi \etal \cite{sarandi2018robust}, we use three occlusion patterns (oriented bars, circles, rectangles) and generate three sets of data, one under each such occlusion pattern (see Fig.~\ref{fig:occlusion} for an illustration). Note that these synthetic occlusions are used only at test time, not during model training.

\paragraph{\textup{\textbf{Baseline architecture evaluation.}}} We first evaluate the efficacy of our proposed regressor design. In Table~\ref{tab:baselineArch}, we compare HKMR's to the currently dominant methods in the encoder-regressor paradigm- HMR \cite{kanazawa2018end} and CMR  \cite{kolotouros2019convolutional}. Note that we use exactly the same encoder (ResNet50) as these methods, so we essentially compare our proposed regressor with their regressors. Here, ``Standard" refers to the standard/existing Human3.6M test set (same as in HMR \cite{kanazawa2018end}), whereas each of the last three columns refers to the Human3.6M test set with the corresponding synthetically generated occlusion pattern. We make several observations from these results. First, on the standard Human3.6M test set, our proposed method outperforms both HMR \cite{kanazawa2018end} and CMR \cite{kolotouros2019convolutional} for both P1 and P2 protocols. For example, the MPJPE of our method for P1 is 71.08 mm, almost 17 mm lower than that of HMR. Similarly, our MPJPE is almost 4 mm lower than CMR in both P1 and P2. Next, these trends hold even under all the three types of occlusions. For example, for ``Bar", our MPJPE is approximately 21 mm and 24 mm lower than HMR for P1 and P2. Finally, our method takes about 0.044 seconds per image, whereas the corresponding numbers for CMR and HMR are 0.062 seconds and 0.009 seconds on a Tesla V100 GPU.

\begin{table}[h!]
  \begin{center}
  \scalebox{0.9}{
    \begin{tabular}{c c cccccccc}
    \toprule
       & \multirow{2}{1.5cm}{\centering \#Param}  & \multicolumn{2}{c}{Standard}  &  \multicolumn{2}{c}{Bar} & \multicolumn{2}{c}{Circle} & \multicolumn{2}{c}{Rectangle}  \\
    \cline{3-10}
       &  &   P1  &  P2  &    P1  &  P2 &    P1  &  P2 &    P1  &  P2 \\
    \midrule
     HMR \cite{kanazawa2018end} & 26.8M & 87.97 & 88.00 & 98.74  & 98.54  & 95.28 & 91.71 & 100.23  & 99.61  \\
     CMR \cite{kolotouros2019convolutional} & 42.7M & 74.70  & 71.90 & 82.99 & 78.85 & 83.50 & 79.24 & 89.01 & 84.73 \\
     \midrule
     \textbf{HKMR}  & 26.2M & \textbf{71.08} &  \textbf{67.74} & \textbf{78.34} & \textbf{74.91} & \textbf{77.60}  & \textbf{71.38}  & \textbf{81.33}  & \textbf{76.79}  \\
    \bottomrule
    \end{tabular}
    }
  \end{center}
  \caption{MPJPE (lower the better) baseline architecture evaluation on Human3.6M.}
    \label{tab:baselineArch}
\end{table}

\begin{figure}
\centering
\includegraphics[height=2.8cm]{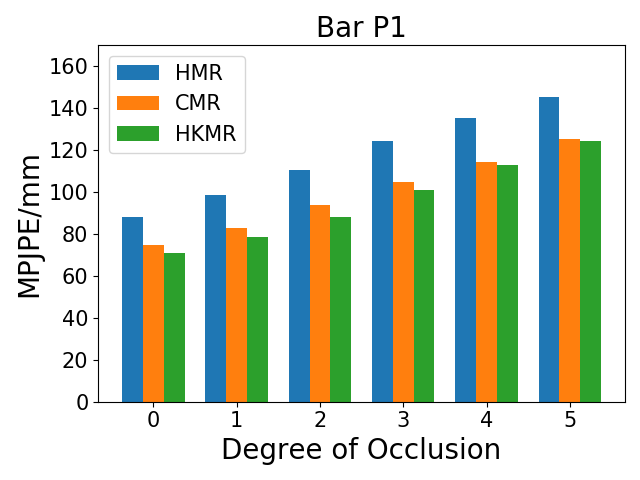}
\includegraphics[height=2.8cm]{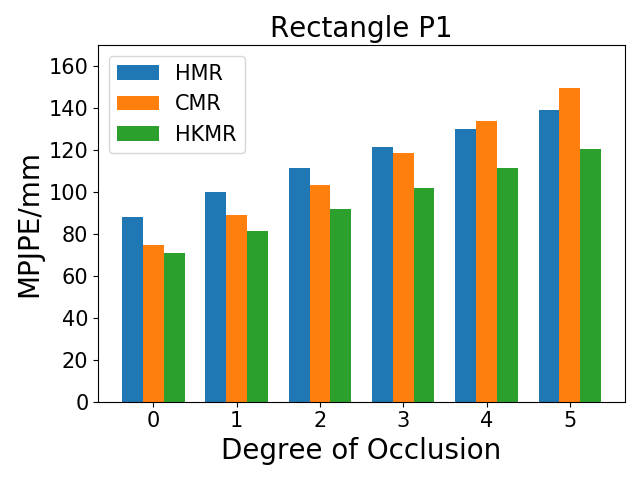}
\includegraphics[height=2.8cm]{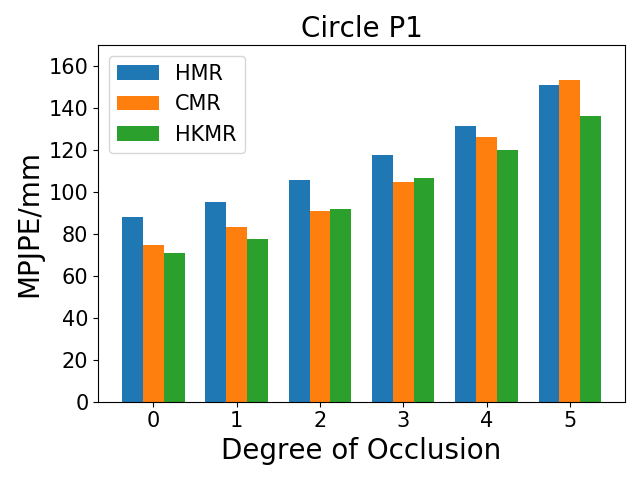}\\
\includegraphics[height=2.8cm]{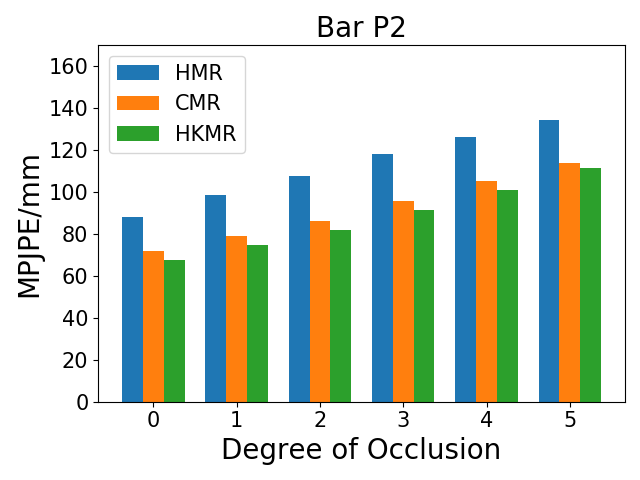}
\includegraphics[height=2.8cm]{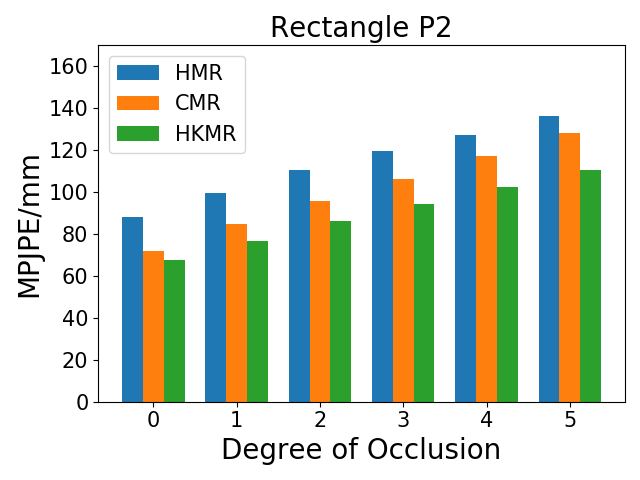}
\includegraphics[height=2.8cm]{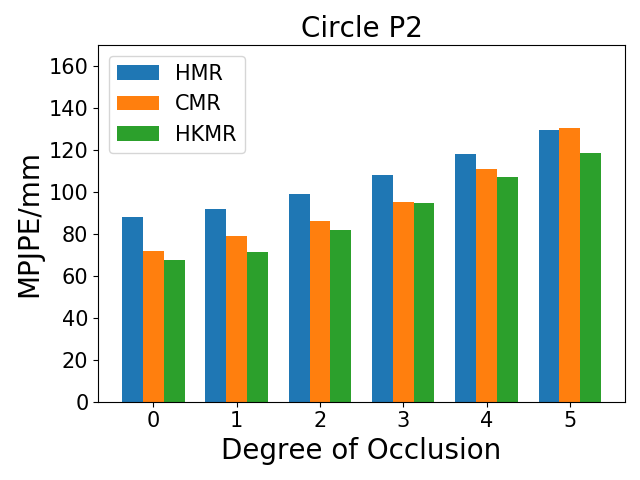}
\caption{Robustness to occlusions: HMR, CMR, and our proposed method.}
\label{fig:occlusionRobustnessBaseline}
\end{figure}

\paragraph{\textup{\textbf{Robustness to degree of occlusions.}}} We next evaluate the robustness of our regressor as we vary the degree of occlusion in the data. For ``Bar", we do this by increasing the width of the bar uniformly from 10 pixels (degree of occlusion or DoC of 1) to 50 pixels (degree of occlusion or DoC of 5). For ``Circle", we increase the radius of the circle from 10 pixels (DoC 1) to 50 pixels (DoC 5). Finally, for ``Rectangle", we increase the area of the rectangle from 3,000 pixels (DoC 1) to 15,000 pixels (DoC 5). In each case, we perform the occlusion on one joint at a time and compute the average MPJPE over all the joints. The results of this experiment are shown in Figure~\ref{fig:occlusionRobustnessBaseline}, where the first row shows the average MPJPE for protocol P1, whereas the second row shows this number for P2. As expected, as the DoC increases, the average MPJPE increases for all three methods. However, this increase is lower for our proposed method, resulting in generally lower average MPJPE values when compared to both HMR and CMR. 

\begin{table}[h!]
  \begin{center}
  \scalebox{0.9}{
    \begin{tabular}{ccccc}
    \toprule
     & \quad No joint hierarchy \quad & \quad  Forward only \quad & \quad  Discriminator \quad & \quad Full model \quad\\
    \midrule
     P1  & 77.10 &  75.99  & 74.21 & \textbf{71.08} \\
     P2  & 74.28 &  72.10 & 71.72 & \textbf{67.74} \\
     \bottomrule
    \end{tabular}}
  \end{center}
  \caption{MPJPE (lower the better) ablation results on Human3.6M.}
  \label{tab:ablationResults}
  \vspace{-2em}
\end{table}

\paragraph{\textup{\textbf{Ablation Study.}}} We next conduct an ablation study to understand the impact of various design considerations in our proposed regressor. The results are shown in Table~\ref{tab:ablationResults}. ``No joint hierarchy" indicates the scenario in our regressor design where each joint prediction is dependent on just the immediately preceding joint, not all the previous joints. ``Forward only" corresponds to conducting only the \textit{forward} pass in the inner-iterative process, not the full forward-backward cycle. ``Discriminator" indicates our full model but using a discriminator \cite{kanazawa2018end} instead of our VAE for the pose prior. ``Full model" indicates our proposed HKMR. As can be seen from the MPJPE values for both P1 and P2, modeling all the joint interdependencies hierarchically as in HKMR is important, with an MPJPE gain of almost 6 mm for both protocols. 
Similarly, there is a 5mm gain when modeling both forward and backward joint dependence, and finally, our full HKMR model with the VAE performs better than the discriminator, resulting in a 3-4 mm MPJPE gain across both protocols. Finally, we also conduct experiments with a varying number of inner and outer iterations (see Table \ref{tab:outer_inner}). One can note that while increasing the number of outer or inner iterations reduces MPJPE, too many iterations actually leads to diminishing returns. In our experiments, we found 3 outer and 4 inner iterations worked best.

\begin{table}[h!]
  \begin{center}
  \scalebox{1}{
    \begin{tabular}{ccccc}
    \toprule
    \textbf{\#Outer Iter}   &   1  &  2  &  3  &  4 \\
    \midrule
     P1     & 93.79 & 80.96 & \textbf{71.08} & 73.16\\
     P2     & 88.61 & 76.41 & \textbf{67.74} & 69.40\\
    \bottomrule
    \end{tabular}
    \begin{tabular}{ccccc}
    \toprule
    \textbf{\#Inner Iter}   &   2  &  4  &  6 \\
    \midrule
     P1     & 83.34 & \textbf{71.08} & 73.61\\
     P2     & 82.94 & \textbf{67.74} & 69.55\\
    \bottomrule
    \end{tabular}
    }
  \end{center}
  \caption{MPJPE (lower the better) ablation results with varying numbers of outer (left) and inner (right) iterations on Human3.6M.}
    \label{tab:outer_inner}
	\vspace{-2em}
\end{table}

\begin{figure}[!h]
\centering
\includegraphics[scale=0.38]{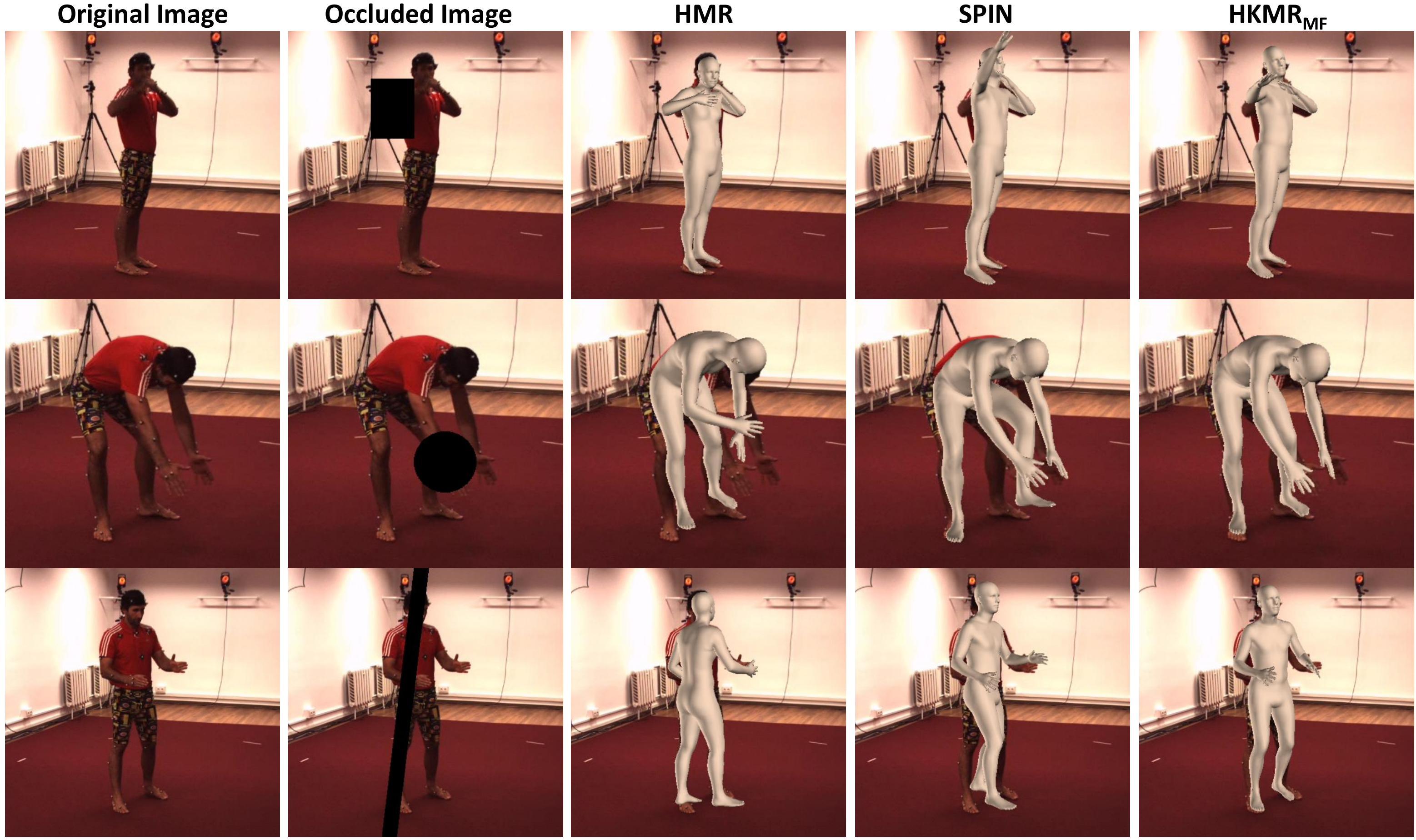}
\caption{Our proposed method results in more accurate fits: right hand in first row, left leg in second row, and root orientation in third row.}
\label{fig:spinComparisonOcc}
\end{figure}

\begin{figure}[th]
\centering
\includegraphics[width=12.3cm]{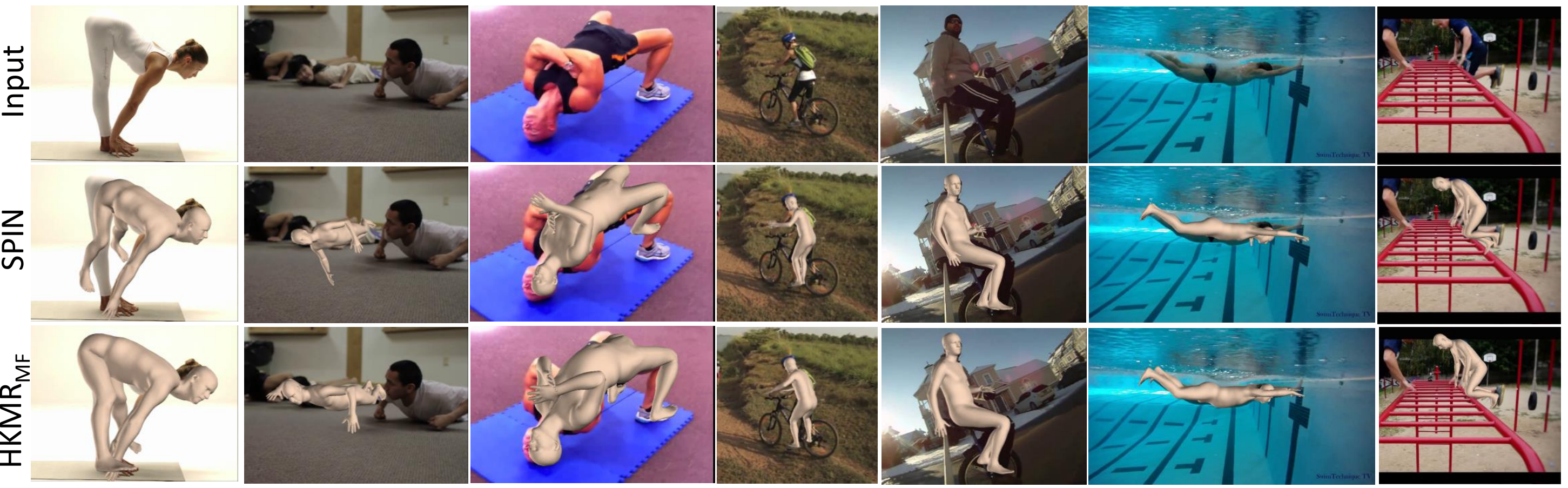}
\caption{Qualitative results on the invisible parts in the MPII validation set.}
\label{fig:MPIIQualitative}
\end{figure}

\paragraph{\textup{\textbf{In-the-loop optimization evaluation.}}} As discussed in Section~\ref{sec:inTheLoopSec}, the in-the-loop optimization techniques follow the encoder-regressor-optimizer paradigm, with SPIN \cite{kolotouros2019learning} the first method to be proposed under this framework. SPIN used the same regressor design as HMR \cite{kanazawa2018end} for predicting the model parameters. Clearly, our proposed method is applicable to this scheme as a simple drop-in replacement of the HMR regressor with our proposed regressor (HKMR$_{MF}$). We present the results of this experiment in Table~\ref{tab:inTheLoop}. As can be noted from the results, our proposed regressor gives lower MPJPE compared to SPIN \cite{kolotouros2019learning} for both P1 and P2. In fact, our MPJPE result of 64.02 mm and 59.62 mm establishes a new state-of-the-art on Human3.6M. Furthermore, we observe similar trends under the three different kinds of occlusions as well. Qualitative results in Figure~\ref{fig:spinComparisonOcc} reinforce this point. For instance, our method results in a better fit in the right hand region (first row) and left leg region (second row) even under occlusions. To further demonstrate the efficacy of our method under occlusions, we also report results on the MPII invisible joints validation dataset \cite{andriluka20142d} in Table~\ref{tab:inTheLoop} (right), where we see substantial gains compared to SPIN for both MPJPE and PCK. Figure~\ref{fig:MPIIQualitative} shows some qualitative results on this data where we see obvious differences - our method results in substantially better model fits even in challenging cases such as those in columns 1-3.

\begin{table}[h!]
  \begin{center}
  \scalebox{0.865}{
    \begin{tabular}{c cc cc cc cc}
    \toprule
      \textbf{Human3.6M}   & \multicolumn{2}{c}{Standard}  &  \multicolumn{2}{c}{Bar} & \multicolumn{2}{c}{Circle} & \multicolumn{2}{c}{Rectangle}  \\
    \cline{2-9}
       MPJPE (mm)$\downarrow$  &   P1  &  P2  &    P1  &  P2 &    P1  &  P2 &    P1  &  P2 \\
    \midrule
     SPIN \cite{kolotouros2019learning} & 65.60 & 62.23 & 74.40 & 68.61 & 74.06 & 67.03 & 77.21 & 70.35  \\
     \textbf{HKMR$_{MF}$}   & \textbf{64.02} &  \textbf{59.62} & \textbf{70.10} & \textbf{64.91} & \textbf{69.60} & \textbf{63.22} & \textbf{70.10} & \textbf{64.91} \\
    \bottomrule
    \end{tabular}
    \begin{tabular}{ccc}
    \toprule
      \textbf{MPII}   &   MPJPE  &  PCK  \\
      \textbf{Invisible} & (pixel)$\downarrow$ & ($\%$)$\uparrow$ \\
    \midrule
     SPIN \cite{kolotouros2019learning}    & 59.52 &  62.16 \\
     \textbf{HKMR$_{MF}$}                   & \textbf{55.56}  & \textbf{66.24} \\
    \bottomrule
    \end{tabular}
    }
  \end{center}
  \caption{Model fitting in-the-loop evaluation on Human3.6M and MPII invisible joints.}
    \label{tab:inTheLoop}
	\vspace{-2em}
\end{table}

\paragraph{\textup{\textbf{Comparison with the state of the art.}}} Finally, we compare the results of our method with competing state-of-the-art methods on the Human3.6M and LSP test set (see Table~\ref{tab:sota}). Note that we follow the same protocol as Kolotouros \cite{kolotouros2019learning} in the evaluation on the LSP dataset. From Table~\ref{tab:sota}, our proposed method gives the highest accuracy and F1-score on the LSP dataset and the lowest MPJPE for both protocols on the Human3.6M dataset, thereby establishing new state-of-the-art results on both these datasets. To further put these numbers in perspective, our method substantially outperforms even competing methods that use extra information, \eg, Arnab \etal \cite{arnab2019exploiting} that uses temporal image sequences as opposed to just one single frame, or DenseRaC \cite{xu2019denserac} that uses UV maps as an additional source of supervision. Finally, even compared to the recently published work of Kocabas \etal \cite{kocabas2019vibe}, our proposed method gives almost 6 mm lower MPJPE. In fact, even Kocabas \etal \cite{kocabas2019vibe} uses extra temporal information as opposed to just one single frame in our case. The strong performance of our method despite these seemingly disadvantageous factors further substantiates the motivation and competitiveness of our technique. As discussed previously, note that all numbers in Table~\ref{tab:sota} are standard MPJPE values without any Procrustes (rigid transformation) post-processing. 

\newcommand{\PreserveBackslash}[1]{\let\temp=\\#1\let\\=\temp}
\newcolumntype{C}[1]{>{\PreserveBackslash\centering}m{#1}}

\begin{table}[h!]
  \begin{center}
  \scalebox{0.9}{
    \begin{tabular}{C{3.15cm}C{.9cm}C{.9cm}C{.9cm}C{.9cm}}
    \toprule
    \multirow{2}{3.15cm}{\centering \textbf{LSP}}  & \multicolumn{2}{c}{FB Seg.}  &  \multicolumn{2}{c}{Part Seg.} \\
    \cline{2-5}
         &   acc.  &  f1  &   acc. &  f1  \\
    \midrule
     Oracle \cite{bogo2016keep}         & 92.17 & \textbf{0.88} & 88.82 & 0.67 \\
     SMPLify \cite{bogo2016keep}                & 91.89 & \textbf{0.88} & 87.71 & 0.64 \\
     SMPLify+\cite{pavlakos2018learning}    & 92.17 & \textbf{0.88} & 88.24 & 0.64 \\
     HMR \cite{kanazawa2018end}                 & 91.67 & 0.87 & 87.12 & 0.60 \\
     CMR \cite{kolotouros2019convolutional}     & 91.46 & 0.87 & 88.69 & 0.66 \\
     TexturePose \cite{pavlakos2019texturepose} & 91.82 & 0.87 & 89.00 & 0.67 \\
     SPIN \cite{kolotouros2019learning}         & 91.83 & 0.87 & 89.41 & 0.68 \\
     \midrule
     \textbf{HKMR$_{MF}$ }                  & \textbf{92.23} & \textbf{0.88} & \textbf{89.59} & \textbf{0.69} \\
    \bottomrule
    \end{tabular}}
    \scalebox{0.9}{
    \begin{tabular}{C{3.15cm}C{1cm}C{1cm}}
    \toprule
      \textbf{Human3.6M}   &   P1  &  P2  \\
    \midrule
     HMR \cite{kanazawa2018end}                  & 87.97 & 88.00 \\
     Arnab \etal \cite{arnab2019exploiting}     & -     &  77.80 \\
     HoloPose \cite{guler2019holopose}           & -     & 64.28 \\
     CMR \cite{kolotouros2019convolutional}      & 74.70  & 71.90 \\
     DaNet \cite{zhang2019danet}                 & -     & 61.50 \\
     DenseRaC \cite{xu2019denserac}              & 76.80  & -\\
     VIBE \cite{kocabas2019vibe}              & -  & 65.60\\
     SPIN \cite{kolotouros2019learning}    & 65.60 &  62.23 \\
     \midrule
     \textbf{HKMR$_{MF}$}                   & \textbf{64.02}  & \textbf{59.62} \\
    \bottomrule
    \end{tabular}}
  \end{center}\caption{Segmentation ($\% \uparrow$) and MPJPE (mm $\downarrow$) on LSP and Human3.6M (``-" indicates result is not reported/unavailable in original papers).}
  \label{tab:sota}
  \vspace{-2em}
\end{table}

\section{Summary}
We presented a new architecture for regressing the pose and shape parameters of a parametric human mesh model that is explicitly informed by the structural knowledge of the model being fit. The proposed new design is quite flexible, which we demonstrated by means of applicability to both the popular encoder-regressor and encoder-regressor-optimizer paradigms for 3D human mesh estimation. By means of extensive experiments on standard benchmark datasets, we demonstrated the efficacy of our proposed new design, establishing new state-of-the-art performance and improved robustness under a wide variety of data occlusions.

\end{document}